\pdfoutput=1
\documentclass[11pt]{article}

\usepackage[final]{acl}

\usepackage{times}
\usepackage{latexsym}
\usepackage{booktabs}
\usepackage{multirow}
\usepackage{colortbl}
\usepackage{makecell}
\usepackage{amssymb}
\usepackage{lipsum}
\usepackage{xcolor}
\usepackage{hyperref}
\usepackage{listings}
\usepackage{mdframed}
\usepackage{fvextra}
\usepackage{tabularx}
\definecolor{codeblue}{rgb}{0.26, 0.44, 0.76}
\definecolor{codeblack}{rgb}{0.0, 0.5, 0.0}
\definecolor{codegray}{rgb}{0.5, 0.5, 0.5}
\definecolor{codepurple}{rgb}{0.58, 0.0, 0.82}
\definecolor{backcolor}{rgb}{0.95, 0.95, 0.92}

\lstdefinestyle{mystyle}{
  language=Python,
  backgroundcolor=\color{backcolor},   
  commentstyle=\color{codeblack},
  keywordstyle=\color{codeblue},
  numberstyle=\tiny\color{codegray},
  stringstyle=\color{codepurple},
  basicstyle=\ttfamily\footnotesize,
  breakatwhitespace=false,         
  breaklines=true,                 
  captionpos=b,                    
  keepspaces=true,                 
  numbers=left,                    
  numbersep=5pt,                  
  showspaces=false,                
  showstringspaces=false,
  showtabs=false,                  
  tabsize=4
}
\lstdefinestyle{promptstyle}{
    basicstyle=\ttfamily\color{black}, % 白色字体
    backgroundcolor=\color{white}, % 黑色背景
    keywordstyle=\color{white}, % 关键字样式也设置为白色
    basicstyle=\ttfamily,
    breaklines=true, % 确保自动换行
    frame=single, % 单一黑色边框
    frameshape={RYRYNYYYY}{yny}{yny}{RYRYNYYYY}, % 定制框架样式
    breakindent = 0pt
}

\newcommand{\name}{Citekit }
\newcommand{\inp}{\textbf{\textsc{Input}} }
\newcommand{\gen}{\textbf{\textsc{Generation Module}} }
\newcommand{\enhance}{\textbf{\textsc{Enhancing Module}} }
\newcommand{\eval}{\textbf{\textsc{Evaluator}} }
\newcommand{\vani}{{\sc{Vanilla}}}
\newcommand{\summ}{{\sc{Summ}}}
\newcommand{\snippet}{{\sc{Snippet}}}
\newcommand{\interact}{{\sc{Interact}}}
\usepackage[T1]{fontenc}
\usepackage{tabularx}
\usepackage[utf8]{inputenc}

\usepackage{microtype}

\usepackage{inconsolata}

\usepackage{graphicx}
\newcommand\tf[1]{\textbf{#1}}
\title{Citekit: A Modular Toolkit for Large Language Model Citation Generation}

\author{Jiajun Shen$^{1,3}$\textsuperscript{\textbf{*}}, 
Tong Zhou$^1$\textsuperscript{\textbf{*}}, 
Yubo Chen$^{1,2}$$^\dag$, 
Kang Liu$^{1,2,4}$\\
$^1$The Key Laboratory of Cognition and Decision Intelligence for Complex Systems \\
Institute of Automation, Chinese Academy of Sciences\\
$^2$School of Artificial Intelligence, University of Chinese Academy of Sciences\\
$^3$University of Chinese Academy of Sciences\\
$^4$Shanghai Artificial Intelligence Laboratory  \\
shenjiajun21@mails.ucas.ac.cn,
tong.zhou@ia.ac.cn\\
\{yubo.chen, kliu\}@nlpr.ia.ac.cn
}

\begin{document}
\maketitle
\let\thefootnote\relax\footnotetext{\textsuperscript{\textbf{*}}These authors contributed equally to this work.}
\let\thefootnote\relax\footnotetext{\textsuperscript{\textbf{†}}Corresponding author.}
\footnote{A demonstration video for Citekit is available at \href{https://youtu.be/KaNICbbmCn0}{https://youtu.be/KaNICbbmCn0}.}
\begin{abstract}

The emerging paradigm of enabling Large Language Models (LLMs) to generate citations in Question-Answering (QA) tasks is lacking in a unified framework to standardize and fairly compare different citation generation methods, leading to difficulties in reproduction and evaluation. Therefore, we introduce Citekit, an open-source and modular toolkit designed to facilitate the implementation and evaluation of existing citation generation methods, while also fostering the development of new approaches to improve citation quality. This tool is highly extensible, allowing users to utilize 4 main modules and 14 components to construct a pipeline, evaluating an existing method or innovative designs. Our experiments with two state-of-the-art LLMs and 11 citation generation baselines demonstrate varying strengths of different modules in answer accuracy and citation quality improvement, as well as the challenge of enhancing granularity. Based on our analysis of the effectiveness of components, we propose a new method, PEEP, obtaining a balanced answer accuracy and citation quality. Citekit is released at \href{https://github.com/SjJ1017/Citekit}{https://github.com/SjJ1017/Citekit}.
\end{abstract}

\section{Introduction}

\label{sec:introduction}

\begin{figure}[t]
\centering
  \includegraphics[width= 1.0\columnwidth]{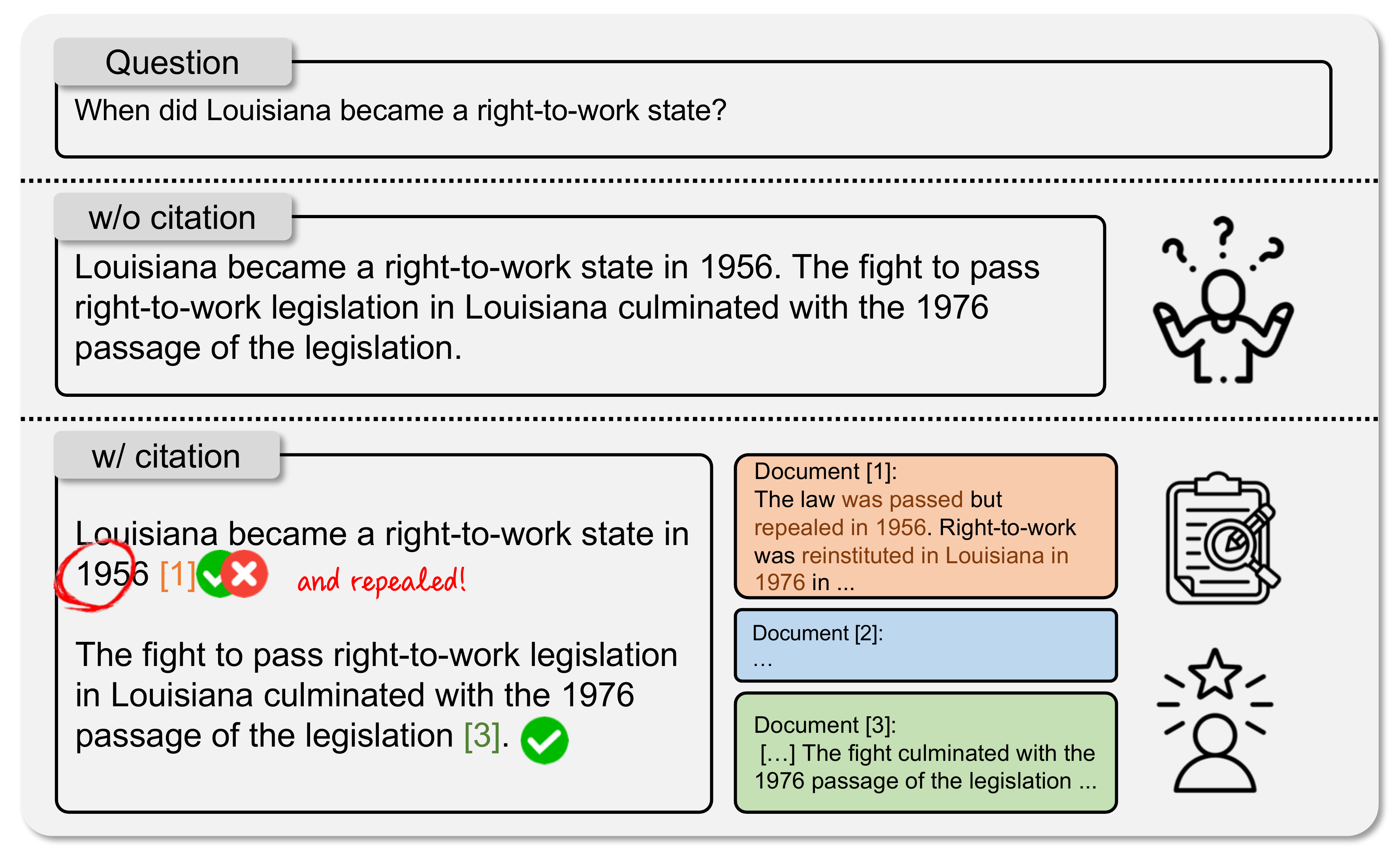}
  \caption{Illustration of Citation Task. An answer without citation makes readers confused about the actual timeline, but if citations are included, they can understand how the details in the answer actually make sense.}
  \label{fig:cite}
\end{figure}

Large Language Models (LLMs) \citep{openai2024gpt4, llama3modelcard} nowadays demonstrate strong performance on Question Answering (QA) \citep{kamalloo2023evaluating} on different scenarios such as Commonsense QA \citep{talmor-etal-2019-commonsenseqa}, long-form QA \citep{stelmakh2023asqa, min-etal-2020-ambigqa} and Multi-hop QA \citep{ho-etal-2020-constructing, yang2018hotpotqadatasetdiverseexplainable}, but they can still inevitably produce hallucinated responses that are non-factual \citep{huang2023survey}, nonsensical or irrelevant to the input\citep{xu2024hallucination}, reflecting the ongoing challenges in ensuring factual accuracy. Given the challenges above, Retrieval Augmented Generation (RAG) \citep{lewis2021retrievalaugmented}, which leverages facts from external unstructured knowledge helps to enhance the reliability of LLMs and can be made more faithful and verifiable by generating citations \citep{gao2024retrievalaugmentedgenerationlargelanguage}. Asking models to generate citations can improve the factual correctness of answers \citep{gao2023enabling}, and the citations that link to the original references will allow readers to easily verify the source of the response, making the answers of models more verifiable and explainable. Figure \ref{fig:cite} shows how citation generation can help users become more assured of the answer. In Figure \ref{fig:cite}, the answer without citation inconsistently states both 1956 and 1976 as the dates for the passage of right-to-work legislation in Louisiana, leading to uncertainty about the actual timeline. If citations are included, readers can scrutinize the reference to understand clearly why there are 2 different dates to the question.

Given the urgent need, ALCE \citep{gao2023enabling} developed some basic methods to enable LLMs to generate citations in QA tasks and propose metrics for evaluating the quality of citations. Following ALCE's contribution, there are other methods that either use training \citep{huang2024training,li2024improving,ye2024effective,huang2024learningfinegrainedgroundedcitations} or construct complicated pipelines to enhance the ability of generative models in citing external documents \citep{,zhang2024verifiable,sun2024verifiabletextgenerationevolving,lee2023reliable,fierro2024learning, qian2024capacitycitationgenerationlarge}. Another category related to citation generation is LLM attribution \citep{jain20231pager,xu2024searchinthechain,gao-etal-2023-rarr,sun2023recitationaugmented, huang2024advancinglargelanguagemodel,cattan2024localizingfactualinconsistenciesattributable, abolghasemi2024evaluationattributionbiasretrievalaugmented}, which refers to the capacity of an LLM to generate and provide evidence \citep{li2023surveylargelanguagemodels}. 

Despite a variety of state-of-the-art methods, there are still two problems:

\textbf{Challenges in reproducibility and improvement:} Different works are distinguished largely on their implementation, making it difficult to reproduce and improve. There are still some works not transparent enough, with inaccessible codes, and different works are realized using different frameworks and in different coding styles, hence the difficulty in generalization and a lack of flexibility. 

\textbf{Need for comprehensive and fair comparisons:} There is a lack of comprehensive and fair horizontal comparisons between various methods. Previous works have not been examined on new LLMs, and since there are now some new metrics like citation granularity, a comprehensive evaluation is needed. Works following ALCE \citep{sun2024verifiabletextgenerationevolving,lee-etal-2024-ask,slobodkin2024attribute} only compare their methods to ALCE baselines. Others \citep{asai2023selfrag,sun2023recitationaugmented} that focus more on answer accuracy are not evaluated on the citation benchmark, which means a lack of horizontal comparisons between SOTAs. 

Given the problems above, a toolkit that unifies different methods is crucial for fast workflow implementation, fair comparisons between various methods, and efficient improvement and innovation. Therefore, we present \name, an open-source, extensible, and user-friendly modular toolkit to construct pipelines for citation generation.

\name offers four different types of modules: \inp, \gen, \enhance, and \eval, which are combined in a pipeline. Input contains automatic components for loading data and making prompts, and is accessible by other modules. \gen is for response generation, where LLM follows the instructions and uses retrieved documents to generate an answer with citations or explicit reference to the documents for further process, featuring a wide range of LLM's supported and adaptive generation modes, it can satisfy different need for various generation task. \enhance contains some components that can assess, retrieve, plan, and edit, and they can be customized for different tasks and even combined into clusters. \eval integrates different metrics to evaluate the output of the pipeline and other new metrics could also be defined and utilized. For training-based methods, a parallel data export component can output evaluation results for supervised learning and Reinforcement Learning (RL). \name also provides significant versatility in customizing new modules to quickly and conveniently realize an improved method. We provide 11 baseline recipes using \name to comprehensively and fairly compare these SOTA methods with the latest models Llama3 and GPT-4o, showing the strength of different modules and remaining challenges. Finally, we learn from the baselines to combine the most efficient planning module, reviser, and simplifier to build our method, PEEP. We achieve a balance in both answer accuracy and citation quality. Our contributions can be summarized as follows:
\begin{itemize}
    \item 
We propose a framework that modularizes citation tasks, with four main modules decomposing citation pipelines into request wrapping, generating, enhancing, and evaluating to unify different methods. The framework contains 14 components and 16 functions to define complicated interconnections between modules and satisfy different needs for citation generation tasks. 
    \item 
We design and complete \name, an easy-to-use and extensible toolkit based on our framework to help reproduce, compare, and combine different methods. We pre-defined 11 recipes to cover 11 citation generation paradigms, all derived from SOTA research.
    \item 
We conduct a comprehensive evaluation and comparison of the existing 11 baselines on 2 SOTA LLMs and propose an improved new method, PEEP, by combining effective components. Our method achieves balance in answer accuracy and citation quality, showing the convenience of verifying new methods.
\end{itemize}

%These instructions are for authors submitting papers to *ACL conferences using \LaTeX. They are not self-contained. All authors must follow the general instructions for *ACL proceedings,\footnote{\url{http://acl-org.github.io/ACLPUB/formatting.html}} and this document contains additional instructions for the \LaTeX{} style files.

%The templates include the \LaTeX{} source of this document (\texttt{acl\_latex.tex}),
%the \LaTeX{} style file used to format it (\texttt{acl.sty}),
%an ACL bibliography style (\texttt{acl\_natbib.bst}),
%an example bibliography (\texttt{custom.bib}),
%and the bibliography for the ACL Anthology (\texttt{anthology.bib}).

\section{System Design}

\begin{figure*}[t]
  \includegraphics[width=\textwidth]{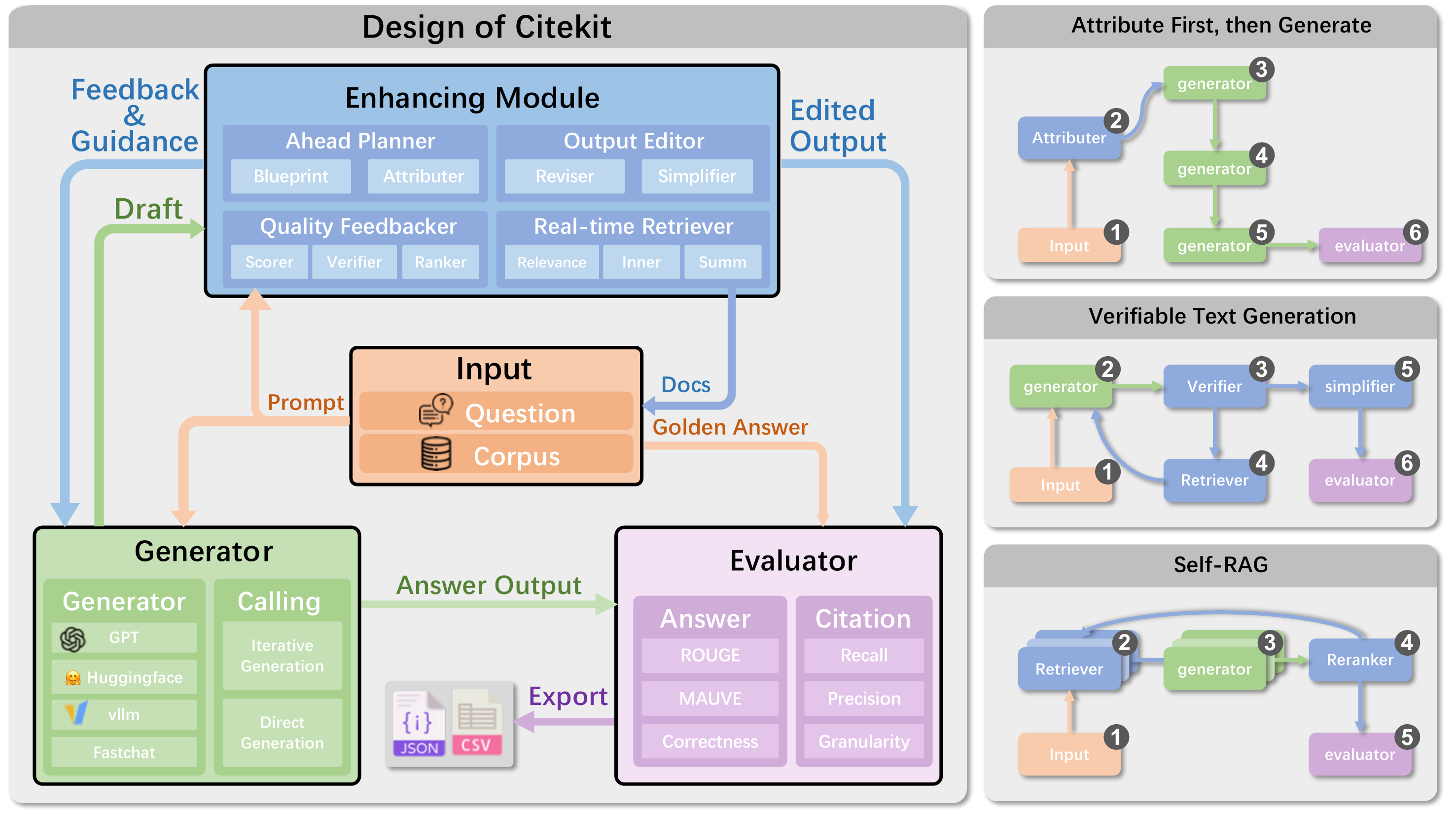}
  \caption{The modular design of \name. On the left, we show four main modules in \name and how they interact with other modules, as well as some predefined components and their abilities; on the right, we illustrate three baseline implementations in our framework and show the data flow during the running of their pipelines}
  \label{fig:design}
\end{figure*}

In this section, we will introduce the design of \name and detail distinct functionalities of different modules mentioned in \S \ref{sec:introduction}, and how they are connected to each other to form an integrated working pipeline of citation generation. We show our design in Figure \ref{fig:design}

\subsection{\inp}
\inp of a citation generation pipeline or a particular module contains requested and retrieved documents.

\textbf{Documents}: In RAG, the input contains some initial documents with knowledge relevant to the question, and the designated documents will be stored for further use and evaluation, each attached automatically with a unique index to trace.

\textbf{Request}: The request of a specific module is among three options: (1) from the user's query, such as questions. (2) from the module itself, like demonstrations and instructions for a task-specific LLM. (3) a dynamic data flow that changes in the process, such as the response from the upstream module.

\subsection{\gen}

\gen contains a Large Language Model for generating content according to according to the requirements. This module, allows users to load a large language model or use some LLM API for generating responses. The input of \gen is a natural language query and the output is a response from LLM. A \gen supports different frameworks, including huggingface, vllm, fastchat, and APIs like openai API to implement the generation according to the need. To fit into different pipelines, \gen can be called either iterative or direct.

\subsection{\enhance}

Works that use one or more external modules to enhance the quality of citations can be classified into four categories: retriever, planner, assessor, and editor, as shown in Table \ref{tab:methods}. They can be used individually or collaboratively, providing sufficient flexibility for the construction of a citation generation pipeline. 

\newcommand{\use}{\cellcolor{blue!17}}
\newcommand{\guse}{\cellcolor{green!17}}
\begin{table}[h]
    \centering
    \renewcommand{\arraystretch}{1.1}
    \resizebox{\linewidth}{!}{
        \begin{tabular}{c|c|c|c|c|c}
            \toprule
            &\textbf{Feedbacker}&\textbf{Retriever}&\textbf{Planner}&\textbf{Editor}\\
            \midrule

           \makecell[c]{ALCE\\Vanilla}& & & & \\
           \hline
           
           \makecell[c]{ALCE\\Rerank}&\use reranker & & &\\
           \hline
           
            \makecell[c]{ALCE\\Interact$^*$}& &\use summary& &\\
           \hline
           
           AnG$^*$& & &\use attributer & \\
           \hline
           
           Bluprint& & &\use blueprint&\\
           \hline
           
           AAR&\use scorer& & &\use reviser\\
           \hline
           
           \makecell[c]{Citation\\Augmented}& &\use relevance& &\\
           \hline
           
           VTG&\use verifier& & & \use simplifier\\
           \hline
           
           recitation& &\use inner& &\\
           \hline
           
           self-RAG$^*$&\use reranker&\use relevance& &\\
           
            \bottomrule

        \end{tabular}
        }
    \caption{
        The usage of different modules and ways of generation in our baselines. $^*$ Methods marked with an asterisk use iterative \gen while others use direct.
    }
    \label{tab:methods}
\end{table}

\subsubsection{Real Time retriever}

A Real-Time Retriever, utilized to retrieve external documents from a corpus, is helpful when LLMs find it difficult to attribute from existing retrieved documents. The input of the retriever is a query and the output contains some retrieved documents or chunks. In the design of \name, documents or chunks returned will be automatically added into the pipeline for later evaluation, with a unique index if needed. The retriever can not only retrieve knowledge by \textit{relevance}, like using bm-25 or dense passage retrieval \citep{karpukhin2020dense} but also get documents in the data store by an \textit{index} or samples documents from LLMs or even the \gen itself (\textit{inner}), as used in Recitation Augmented baseline asking LLMs to recite documents from training data.

\subsubsection{Ahead Planner}

Planners will process the query and documents in advance before it is sent to LLMs for generation. Taking the query and relevant documents as input, An ahead planner such as \textit{blueprint} modules and \textit{attributers} generate guidance that the \gen can follow to improve the citation quality. The generated information like plans or attributing routes serve to help \gen better understand and extract knowledge, while also making the answer more traceable.

\subsubsection{Quality Feedbacker}

A Feedbacker, once defined and plugged into the pipeline, can automatically evaluate the initial answer in the process to guide the modules to generate a better response. The input contains the initial answer, and the output of a feedbacker may be a quantitative value (\textit{scorer}), or the answer with the highest score by some pre-defined metrics (\textit{reranker}). A special type of feedbacker, \textit{verifier}, can also output the exact input but present a Boole value True or False for distinguished further process.

\subsubsection{Output Editor}

An output editor can modify the response for a better citation or answer quality. The input of an editor contains an 
answer and the output is a new answer edited using information from the data storage or the other feedback from the input. It can either \textit{revise} the answer, like correct the factual problems in the answer, or modify the citation, including simplifying it to make it more precise (\textit{simplifier}).

\subsubsection{Extensibility}
In addition to the predefined components in \enhance, researchers can also create a new component by just setting a corresponding prompt template and a calling function that demonstrates the logic of execution. Any module that is inherited from the base class will have the ability to be connected to the pipeline and send output to the target module so the new module can be easily plugged into the pipeline.

\subsection{\eval}

\eval is a module that evaluates or scores the output. If the \eval is plugged in, the output of the pipeline will be automatically sent to it, and other information such as reference answers will also be passed into it for evaluation. \eval have access to the initial corpus as well as the newly retrieved documents during the whole process. Finally, a result will be returned by the \eval.

There are some predefined metrics that can be set easily, such as ROUGE for answer quality and MAUVE for fluency, citation precision and recall in ALCE benchmark, a citation precision and recall metric with granularity for citation quality, and dataset-specific metrics for answer correctness (e.g. STR-EM for ASQA, claims for ELI5).

Manually defined other metrics are also possible once the evaluation function and the specific data in the pipeline for evaluation are defined, allowing users to implant an existing metric into a pipeline or a new one.

\subsection{Pipeline}
A pipeline is a class for managing data flow and organizing different modules and corresponding components. It serves as a runner to start the citation generation process and control the input and output of modules contained in itself. 

For data management, input (e.g. question to be answered, ground truth for evaluation) and documents retrieved are stored respectively as dictionaries with keys and values in the pipeline, and they are both accessible by modules. 

For module organization, modules that are connected to the pipeline can take an input and generate an output, and the output will be sent to target a module by predefined conditions.

\name offers more flexible options for more complicated pipelines. For instance, multiple responses can be sent to the next module in parallel for independent processes. They can also be sent iterative for sequential needs. Besides, modules that simply form a sequence can be connected in order and be used like a complete module.

The pipeline is also extensible, as users can plug in different modules. Different ways of connection will make the pipeline a sequence, a loop, a tree, or other structures.

\section{Usage}
\subsection{Realization of SOTA method.} To construct a citation generation pipeline with \name, users can use predefined recipes to define some preset modules and combine them. For Attribute First, then Generate pipeline, users can simply use a list to indicate the interconnection and the last module will output the answer. 

To run the entire pipeline, the user should specify certain entries from the dataset as input, and designate document entries as the initially stored documents. As shown in Figure \ref{fig:code1}, we use several lines of code to complete the method quickly.

\begin{figure}[!thp]
\centering
\begin{minipage}{0.96\linewidth}
\begin{lstlisting}[language=Python]
from Citekit import Enhance, Generator
...
dataset = PromptDataset("asqa.json")
evaluator = DefaultEvaluator()
attribute = Enhance.Attributing()
answer = Generator.LLM(
    output = 'prefix',
    iterable=True)

pipeline = Pipeline(
    sequence = [attribute, answer], 
    evaluator = evaluator, 
    dataset = dataset)

pipeline.run_on_dataset(
    data = ['question', 'docs'], 
    init_docs='docs')

\end{lstlisting}
\end{minipage} 
\caption{An example to define an Attribute First, then Generate pipeline. An attributer will highlight some spans and cluster them, and each cluster will be used to generate an answer sentence.}
\vspace{-0.3cm}
\label{fig:code1} 
\end{figure}

\subsection{Customized pipeline modification.} The Attribute First, then Generate design uses only an ahead planner. If we want to extend the pipeline by plugging in a verifier to ask the \gen to regenerate with new retrieved documents if the statement is not entailed by the documents. We will add a loop after the initial output to \gen. Figure \ref{fig:code2} shows an example of the efficiency of improving an existing method.

\begin{figure}[!thp]
\centering
\begin{minipage}{0.96\linewidth}
\begin{lstlisting}[language=Python]
...
retriever = EnhanceModule.Retriever(
    target = answer)
verifier = EnhanceModule.Verifier(
    target_false = retriever)
answer.set_target(verifier)

pipeline.run_on_dataset(
    data = ['question', 'docs'], 
    init_docs='docs')
\end{lstlisting}
\end{minipage} 
\caption{An example to plug in new modules and recreate a new pipeline.}
\vspace{-0.3cm}
\label{fig:code2} 
\end{figure}

\renewcommand{\arraystretch}{1.0} % 1.5 倍行距

\begin{table*}[h]
    \centering
    \small
        \begin{tabular}{l|ccccccccc}
            \toprule
             &&\tf{Fluency}& \tf{Correct.} & \multicolumn{4}{c}{\tf{Citation} }\\
            \cmidrule(lr){3-3} \cmidrule(lr){4-4} \cmidrule(lr){5-8}
           &Model& (MAUVE) & (EM Rec.) & Rec. & Prec.& App. & Gran. & ROUGE-L & Length \\
            \midrule[1.5pt]

            \multirow{2}{*}{\tf{\shortstack[l]{ALCE\\  \vani}}}&llama3-8B       &66.8&40.5&47.2&53.8&80.5&22.5&28.6&72.0\\
            &GPT-4o            &72.3&41.0&59.5&61.3&70.8&19.3&32.4&41.6\\
            
            \cline{1-10}\\[-3mm]

                        \multirow{2}{*}{\tf{\shortstack[l]{ALCE \\ \summ}}}&llama3-8B            &80.1&40.6&59.5&66.2&80.6&59.7&27.7&69.4\\
           &GPT-4o               &72.3&42.0&59.6&61.4&82.6&54.5&32.5&41.6\\

            \cline{1-10}\\[-3mm]
           
                        \multirow{2}{*}{\tf{\shortstack[l]{ALCE \\ \snippet}}}&llama3-8B      &69.2&38.9&56.7&60.9&81.8&65.6&27.1&65.3\\
            &GPT-4o         &79.7&37.3&77.0&66.8&85.6&58.3&30.2&26.5\\

                        \cline{1-10}\\[-3mm]
                        \multirow{2}{*}{\tf{\shortstack[l]{ALCE\\ \interact}}}&llama3-8B        &68.0&30.3&30.6&56.1&84.1&17.2&21.5&56.6\\
            &GPT-4o           &72.6&39.9&41.2&45.0&    72.0&12.0&30.4&67.3\\

            \midrule[1.5pt]

            \multirow{2}{*}{\tf{\shortstack[l]{Attribute, \\ then Generate}}}&llama3-8B                 &70.2&38.9&49.2&42.7&78.0&22.8&27.9&89.3\\
            &GPT-4o                    &75.5&41.6&63.4&42.7&87.0&19.2&24.8&61.2\\
            
             \cline{1-10}\\[-3mm]
            \multirow{2}{*}{\tf{AAR}}&llama3-8B                 &69.4&38.9&37.8&47.8&74.1&28.1&27.0&122.8\\
            &GPT-4o                    &72.2&46.0&52.4&58.7&77.8&20.9&31.5&59.0 \\
            
             \cline{1-10}\\[-3mm]
            \multirow{2}{*}{\tf{\shortstack[l]{Citation\\ Enhanced}}}&llama3-8B      &59.2&31.0&30.9&40.8&54.0&27.2&24.8&48.7\\
            &GPT-4o         &65.3&41.3&49.8&52.8&55.3&27.0&29.6&40.6\\
            
             \cline{1-10}\\[-3mm]
            \multirow{2}{*}{\tf{VTG}}&llama3-8B&74.9&41.2&73.4&73.1&87.3&27.0&42.4&45.3\\
                                     &GPT-4o       &75.1&42.3&83.0&82.5&88.4&29.3&39.3&45.3\\
            
             \cline{1-10}\\[-3mm]
            \multirow{2}{*}{\tf{Blueprint}}&llama3-8B             &70.0&40.8&68.5&71.3&87.5&22.5&31.2&75.8\\
            &GPT-4o                &78.2&41.2&68.5&83.0&83.6&19.8&27.2&75.8\\

            \midrule[1.5pt]

            \multirow{2}{*}{\tf{\shortstack[l]{Recitation \\ Augmented}}}&llama3-8B             &61.2&33.6&47.6&55.0&62.5&14.4&34.5&129\\
            &GPT-4o$^*$           &/&/&/&/&/&/&/&/\\

                         \cline{1-10}\\[-3mm]
            \multirow{2}{*}{\tf{Self-RAG}}&llama3-8B             &68.4&35.7&82.7&80.2&88.3&28.3&27.1&52.2\\
            &GPT-4o                &70.7&37.9&81.5&83.25&84.6&26.4&27.9&40.7\\
                         \cline{1-10}\\[-3mm]
            \multirow{2}{*}{\tf{PEEP}}
            &llama3-8B&68.3&38.3&78.4&79.3&82.0&66.4&24.6&59.0\\
            &GPT-4o   &75.9&41.9&83.2&85.0&86.2&65.2&28.1&66.7\\
            \bottomrule
        \end{tabular}
    \caption{
        ASQA results. $^*$In recitation-augmented baseline, we only use Llama3-8B-Instruct because we found GPT-4o is too reluctant to recite verbatim documents in training data.%
    }
    \label{tab:asqa_full}
\end{table*}

\section{Evaluation}

\subsection{Baselines and metrics}
We evaluate 11 baselines in total using the state-of-the-art open-source and closed-source LLMs, GPT-4o \citep{openai2024gpt4} and Llama3-8B-Instruct \citep{llama3modelcard} on ASQA dataset. \textbf{ALCE \vani}, \textbf{\snippet}, and \textbf{\summ} directly prompt the LLM to generate citations using full documents, snippets, and summaries respectively. \textbf{ALCE \interact} \citep{gao2023enabling} uses document summaries and interactively provides full documents. \textbf{AAR} \cite{lee-etal-2024-ask} asked the LLM to revise the answer, while \textbf{VTG} \cite{sun2024verifiabletextgenerationevolving} will verify the answer and retrieve more supplementary documents for regeneration. \textbf{Citation Enhanced} \citep{li2024citationenhanced} method retrieves documents after generation, and \textbf{Recitation Augmented} \citep{sun2023recitationaugmented} sample documents from pre-training data. \textbf{Attribute First, then Generate} \citep{slobodkin2024attribute} and \textbf{Blueprint} \cite{fierro2024learning} provides some attributing spans or questions to guide the generation. For \textbf{self-RAG} \citep{asai2023selfrag}, we use our prompt version instead of a trained model to retrieve documents and generate sentence-by-sentence. 
We use metrics from ALCE for evaluation, including fluency, correctness, rouge, citation recall, and precision. We also evaluate the appropriate citation rate and the citation granularity. 
\subsection{Our method}
\begin{figure}[!thp]
\centering
  \includegraphics[width=1.0\columnwidth]{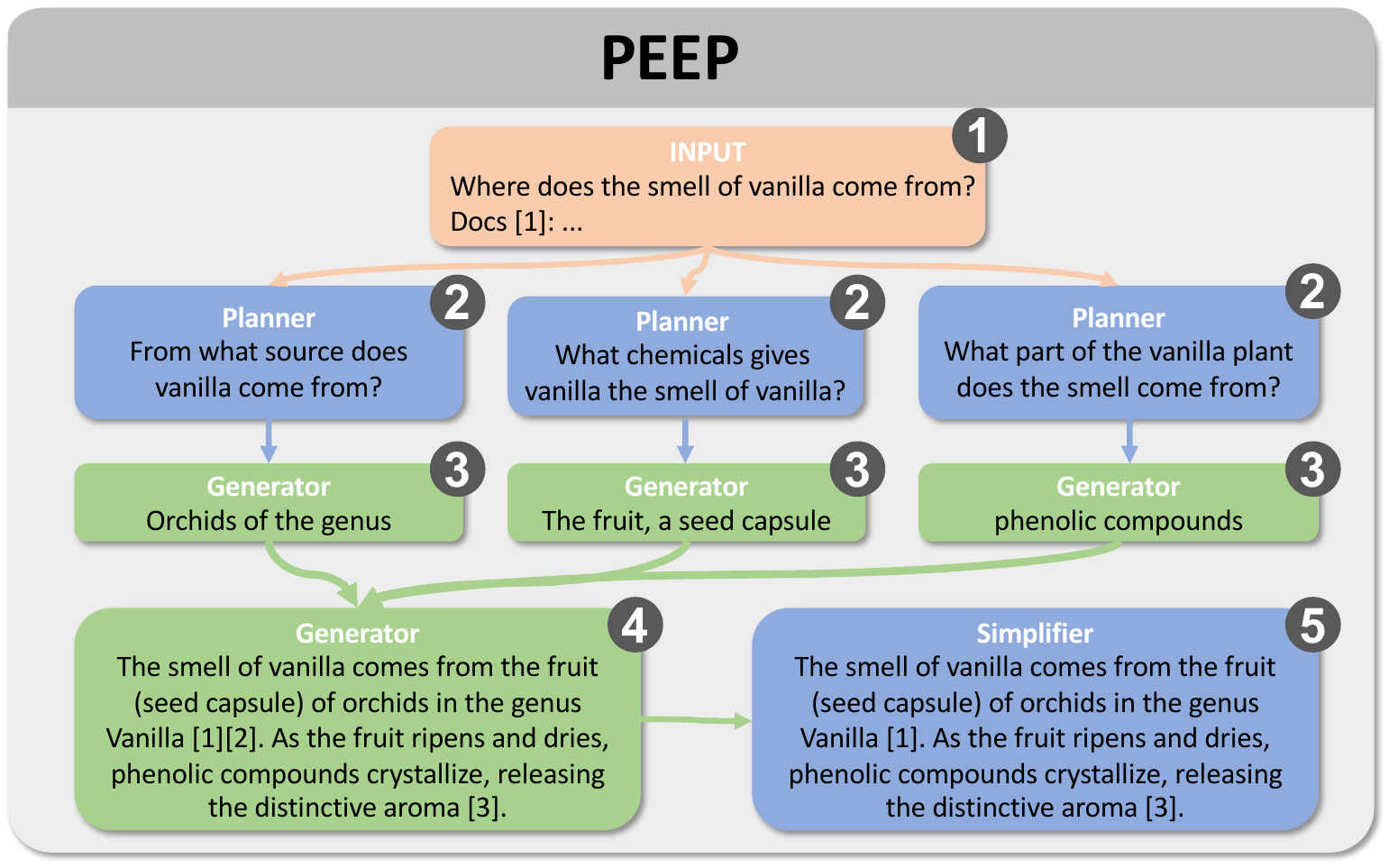}
  \caption{Design of PEEP. We show an example of generating a comprehensive answer for an ASQA question.}
  \label{fig:peep}
\end{figure}

We build our new pipeline, \textbf{PEEP}, combining the most efficient modules. We use a \textbf{P}lanner to decompose the question in two as much as atomic questions \citep{yan2024atomicfactdecompositionhelps}, and \textbf{two E}ditors (reviser and simplifier) to combine the answers for each atomic question and simplify the citation afterward. We use \textbf{P}arallel generation to get answers with sufficient diversity, as shown in Figure \ref{fig:peep}.

\subsection{Settings}
We set max generated tokens to 500 to avoid too long answers and use \texttt{\textbackslash n} as stop token. For Llama3-8B-Instruct, we use the model from huggingface and set the temperature to 0.5. and other configurations by default. For GPT-4o, we use the openai API. During our experiment, we used the same prompt for the two models.

For retrieving documents relevant to the query, we use 5 documents by default. However, for ALCE \summ, ALCE \snippet, and ALCE \interact, we use 10 documents as they show the short summaries and snippets from the documents. Citation Augmented and self-RAG use real-time retrievers instead of a fixed number of document inputs, and we configured our retrievers to return the top-1 document at a time.

For evaluation of citation quality, we adopt a \href{https://huggingface.co/google/t5_xxl_true_nli_mixture}{TRUE model} \citep{honovich-etal-2022-true-evaluating} to verify if the cited documents could entail the generated statement.
\subsection{Results and analysis}
We show the full results on ASQA dataset in Figure \ref{tab:asqa_full}. We discuss the main results from the experiments below.

\textbf{A more advanced model performs better.} GPT-4o is generally better than Llama3-8B-Instruct on both citation quality and answer correctness. The performance of citation generation can benefit from the enhanced capabilities of the base model. 

\textbf{Ahead planner can enhance the answer.} A planner can enhance the answer on correctness, especially for a more powerful model. GPT-4o is more likely to achieve a better performance via planning.

\textbf{Output editor can significantly improve the citation quality.}
An output editor can improve citation recall and precision, as well as the correctness of citation prediction. While the ALCE Vanilla presents a citation precision and recall at about 50, a reranker in self-RAG can make the precision and recall achieve 80.

\textbf{Enhancing the granularity of citations is still a challenge.}
Nearly all the baselines presented cite the full documents, resulting in a relatively low granularity. Citation generation based on summary and extraction(ALCE \summ and ALCE \snippet) can cite only a snippet or a summary from the document, but it risks a loss of correctness.

\textbf{LLMs can cite internal knowledge better}
Despite the significant loss of answer quality and correctness, Llama3-8B-Instruct demonstrates better citation quality both on recall and precision when citing the documents sampled from itself, compared to the ALCE-Vanilla baseline that uses external knowledge, demonstrating the prospects of research in this area.

\textbf{Considering both citation quality and answer correctness remains a challenge} 
Our method significantly improves the overall quality of citations while only slightly sacrificing accuracy. It achieves the best balance among all the methods tested. However, the answer accuracy is still lower than the highest-performing method by 7.3 points. Additionally, the citation recall and precision barely exceed 80\%. In practical applications, it is still difficult to gain readers' trust. We believe that thanks to the modularity and extensibility of \name, this issue will gradually be resolved.

\section{Conclusion}
To unify different methods for LLM citation generation and to conduct a comprehensive and fair comparison of existing methods, we propose \name a user-friendly, modular, and extensible toolkit. We also present an instance to demonstrate the application of the toolkit, showing the usability and versatility of realizing citation generation pipelines. We conducted experiments on 11 different baselines and found that different modules excel in improving either the answer accuracy or the citation quality, and our approach achieves the best balance between answer accuracy and citation quality. However, generating a citation in fine-grained granularity is still challenging. 

\section{Limitation}
There are still areas for improvement in our evaluation. (1) We only conduct our experiment on two LLMs, GPT-4o and Llama3-8B-Instruct. For other models, especially smaller ones, whether the different methods would still be effective in improving the performance of citation tasks is unknown. (2) Existing datasets are not designed for citation tasks. For instance, they do not take into account appropriate citation generation based on needs. Building datasets that reflect real citation generation scenarios remains an open problem.
\bibliography{custom}

\clearpage

\appendix

\section{Implementation Details}
\label{sec:appendix1}
In this section, we describe the implementation details for different baselines. For other baselines, we follow the original prompts and the structure they provided, but for Blueprint and self-RAG, we use In-Context-Learning (ICL) instead of a trained model to complete the sub-task in their design.

\subsection{Blueprint Model}
For the Blueprint Model, we use the abstractive model to produce general-purpose questions: the paragraph is the input and the question is the output. We use prompts to make LLMs generate questions. ALCE provides question-answer pairs for ASQA dataset, and in each pair the sub-question shows an aspect of answering the final question. We use these pairs to complete a 2-shot prompt for ICL. For answer generation, we adjust the ALCE prompt to make LLMs answer all the subquestions. We show our prompts in Figure \ref{fig:blueprint1}, \ref{fig:blueprint2}

\subsection{Prompt self-RAG}
\label{selfrag appendix}
As for Llama3-8B and GPT-4o, there is no trained version for self-RAG, we use prompt to make the LLM retrieve documents and generate, then use an NLI model to evaluate if the document is supportive and the answer is useful, respectively in 3 segments. A reranker will find the best segment and the sentence is add to the answer. Similar to Attribute First, then Generate, We use generated sentences as prefix to complement the sentence-by-sentence iterative generation. We show our prompts in Figure \ref{fig:selfrag1}, \ref{fig:selfrag2}

\section{Case Study}
\label{sec:appendix2}
\subsection{Answer quality improvement}

\DefineVerbatimEnvironment{CustomVerbatim}{Verbatim}{
  commandchars=\\\{\},
  fontsize=\small,
    formatcom=\color{gray}
}

  We will describe how the answer quality could be improved by using a planner. In ASQA, the question is to some extent ambiguous and requires multiple short answers to cover different aspects. As in Figure \ref{fig:answer improve}, a planner can give guidance by a blueprint or properly highlighting and clustering useful spans. 
\begin{figure}[!thp]
    \centering
    \begin{mdframed}[innerrightmargin=20pt]
  \begin{CustomVerbatim}
\textcolor{black}{Question: Who was the 
programmer of MS-DOS}
\textcolor{black}{operating system?}
  
\textcolor{red}{ALCE-Vanilla:}
\textcolor{blue}{Answer:}
\textbf{\textcolor{black}{Tim Paterson}} was the programmer of MS-DOS 
operating system [1]. He developed 86-DOS, 
which later formed the basis of MS-DOS [1]. 
MS-DOS was a renamed form of 86-DOS, written 
by Tim Paterson [3].

\textcolor{red}{Attribute First, then Generate:}
\textcolor{blue}{Attributer:}
1.
Document[1]: 
...MS-DOS was a renamed form of 86-DOS owned 
by Seattle Computer Products, written by Tim 
Paterson...

2.
Document[1]: 
Paterson was educated in ...
Document[4]: 
Microsoft purchased 86-DOS, allegedly for 
\$50,000 ...
\textcolor{blue}{Answer:}
\textbf{\textcolor{black}{Tim Paterson}} is the programmer who developed
the MS-DOS operating system, which was 
initially named 86-DOS [1]. \textbf{\textcolor{black}{Tim Paterson}}, the
developer of 86-DOS, is credited with creating
the operating system, which was later renamed 
MS-DOS, and was purchased by \textbf{\textcolor{black}{Microsoft}} for 
\$50,000 [1][4]. 

\textcolor{red}{Attribute First, then Generate:}
\textcolor{blue}{questions:}
Who was the programmer of 86-dos operating 
system? Which company was the programmer of 
MS-DOS operating system? ...

\textcolor{blue}{Answer:}
\textbf{\textcolor{black}{Tim Paterson}} created 86-DOS, which later 
formed the basis of MS-DOS [1]. MS-DOS is an 
operating system for x86-based personal compu-
ters mostly developed by \textbf{\textcolor{black}{Microsoft}}[4].
   \end{CustomVerbatim}
       \end{mdframed}
    \caption{Answer Improvement}
    \label{fig:answer improve}

\end{figure}

\subsection{Citation quality improvement}
Although we ask the model to cite a minimum set of documents, LLMs still tend to overcite. Since the ASQA dataset contains rare multi-hop questions, most of the statements only need one document as a citation. An editor, such as a simplifier can remove redundant citations. Figure \ref{fig:citation improve} shows the answer before and after simplification in VTG.

\begin{figure}[!thp]
    \centering
    \begin{mdframed}[innerrightmargin=20pt]
  \begin{CustomVerbatim}
\textcolor{black}{Question: When was the first Apple i phone} 
\textcolor{black}{made?}

\textcolor{black}{Documents: } 
\textcolor{black}{Document[1]: ...was later released in the }
\textcolor{black}{United States on June 29, 2007...}
\textcolor{black}{Document[2]: ...the first iPhone would be }
\textcolor{black}{released later that year. On June 29, 2007...}
  
\textcolor{red}{VTG before removal:}
\textcolor{blue}{Answer:}
The first Apple iPhone was released on June
29, 2007 \textbf{\textcolor{black}{[1][2]}}.

\textcolor{red}{VTG after removal:}
\textcolor{blue}{Answer:}
The first Apple iPhone was released on June 
29, 2007 \textbf{\textcolor{black}{[2]}}.
   \end{CustomVerbatim}
       \end{mdframed}
    \caption{Citation Improvement}
    \label{fig:citation improve}

\end{figure}

\subsection{Granularity improvement}
To improve granularity, the answer should cite the minimum number of spans from the documents. Most of the methods use document-level citation, and in our metrics of granularity, we assume all the spans in one document are cited for document-level citation. In ALCE-\snippet, LLM only cites a snippet from the document, hence a high score of granularity. Figure \ref{fig:granularity Improve} shows how ALCE \vani works to cite a span, not a document.

\begin{figure}[!thp]
    \centering
    \begin{mdframed}[innerrightmargin=20pt]
  \begin{CustomVerbatim}
\textcolor{blue}{Documents: }
...
Document [3]: (Title:The Sound of Silence)...
Notes Bibliography The Sound of Silence \textcolor{black}{"The}
\textcolor{black}{Sound of Silence", originally "The Sounds of }
\textcolor{black}{Silence", is a song by the American music duo }
\textcolor{black}{Simon & Garfunkel}. The song was written by...
...

\textcolor{blue}{Provided Snippets: }
...
\textcolor{black}{Document [3]: (Title:The Sound of Silence)"The}
\textcolor{black}{Sound of Silence", originally "The Sounds of }
\textcolor{black}{Silence", is a song by the American music duo} 
\textcolor{black}{Simon & Garfunkel}.
...

\textcolor{blue}{Answer:}
\textcolor{black}{... The song was originally titled "The Sounds }
\textcolor{black}{of Silence" [3]. ...}
   \end{CustomVerbatim}
       \end{mdframed}
    \caption{Granularity Improvement}
    \label{fig:granularity Improve}

\end{figure}

\clearpage
\onecolumn
In figure \ref{fig:answers}, we show an example of all the answers generated by different methods. 

\begin{figure*}[!thp]
    \centering
\noindent 
\begin{tabularx}{\textwidth}{|l|X|} 
\hline
 & Question: When did toronto host the mlb all-star game? \\ \hline
ALCE \vani & The 1991 Major League Baseball All-Star Game was hosted by Toronto on July 9, 1991, at SkyDome [1]. \\ \hline 
ALCE \summ & Toronto hosted the MLB All-Star Game in 1991, when the 1991 MLB All-Star Game was held on July 9, 1991, at SkyDome in Toronto [1]. It did not host the game in any other year mentioned in the provided search results [3][4] \\ \hline
ALCE \snippet & Toronto hosted the MLB All-Star Game in 1991 and 2007, as the game was held on July 9, 1991, at SkyDome in Toronto [1] and on July 10, 2007, in Toronto [3]. \\ \hline

ALCE \interact & The MLB All-Star Game was held on July 10, 2007, in Toronto [3].   \\ \hline

Attribute, Generate &  The MLB All-Star Game was hosted in Toronto on July 9, 1991, at SkyDome, the home of the Toronto Blue Jays [1].The game in Toronto in 1991 saw the American League face off against the National League at the home of the Toronto Blue Jays.The MLB All-Star Game in Toronto was hosted on July 13, 1993, at Camden Yards in Baltimore with Cito Gaston managing the American League squad for the first time [4].\\ \hline

AAR &  Toronto hosted the Major League Baseball All-Star Game on July 9, 1991, at SkyDome, the home of the Toronto Blue Jays [1]. \\ \hline
Citation Enhanced & Toronto hosted the Major League Baseball (MLB) All-Star Game in 1989[1]. 1989[2].  The 2015 Major League Baseball All-Star Game was held in Toronto, Ontario, Canada, on July 14, 2015. [3]  \\ \hline
VTG &  The 1991 MLB All-Star Game was held on July 9, 1991, at SkyDome in Toronto, Canada [1]. It was the second time that the game was played outside the United States, after the 1982 game in Montreal, Quebec, Canada [2]. \\ \hline

Blueprint &  Toronto hosted the MLB All-Star Game in 1991 [1]. The first All-Star Game played outside the United States was in 1982, held at Olympic Stadium in Montreal, Quebec, Canada [2]. The MLB All-Star Game last held in New York City in 2008, when the old Yankee Stadium hosted it [5]. \\ \hline
Recitation Augmented & Toronto has hosted the MLB All-Star Game three times: in 1989, 2002, and 2016. The 1989 game was held on July 11, 1989, at SkyDome (now known as Rogers Centre), and the American League won 5-3 [1]. The 2016 game was held on July 12, 2016, at Rogers Centre, and the American League won 4-2 [1]. Toronto has hosted the MLB All-Star Game a total of three times [1].  \\ \hline
self-RAG &  Toronto hosted the 1991 Major League Baseball All-Star Game at SkyDome on July 9, 1991[1]. Toronto hosted the MLB All-Star Game in 2016, but the provided search result does not mention this specific event, only the 2007 game held at AT\&T Park in San Francisco[4]. \\ \hline

\end{tabularx}

    \caption{Example of final answers}
    \label{fig:answers}
\end{figure*}

\begin{figure*}
    \centering
    
  \begin{Verbatim}[frame=single, framesep=2mm, fontsize=\small, breaklines=true]
In this task, you should write no more than four subquestions according to the given documents and a question. Ensure that each subquestion can be respond by reading the documents, and is related to the question. Write then in only one paragraph.

Question: Who is the original artist of sound of silence?

Document [1]: Sounds of Silence is the second studio album by Simon & Garfunkel, released on January 17, 1966. The album's title is a slight modification of the title of the duo's first major hit, "The Sound of Silence", which originally was released as "The Sounds of Silence". The song had earlier been released in an acoustic version on the album "Wednesday Morning, 3 A.M.", and later on the soundtrack to the movie "The Graduate". 
Document [2]:  Sound of Silence" is a song performed by Australian recording artist Dami Im. Written by Anthony Egizii and David Musumeci of DNA Songs, it is best known as Australia's entry at the Eurovision Song Contest 2016 which was held in Stockholm, Sweden, where it finished 2nd, receiving a total of 511 points.
Document [3]: Simon & Garfunkel Simon & Garfunkel were an American folk rock duo consisting of singer-songwriter Paul Simon and singer Art Garfunkel. They were one of the bestselling music groups of the 1960s and became counterculture icons of the decade's social revolution, alongside artists such as the Beatles, the Beach Boys, and Bob Dylan. Their biggest hits\u2014including "The Sound of Silence" (1964), "Mrs. Robinson" (1968)

Sub-questions:
Who is the original artist of sound of silence, the album? Who is the original artist of sound of silence, the song, released in 2016? Who is the original artist of sound of silence, the song, released in 1964?"


In this task, you should write no more than four subquestions according to the given documents and a question. Ensure that each subquestion can be respond by reading the documents, and is related to the question. Write then in only one paragraph.

Question: ...

Document[1]: ...
Document[2]: ...
Document[3]: ...

Sub-questions: 
        
   \end{Verbatim}

    \caption{Prompt for question generation of Blueprint}
    \label{fig:blueprint1}
\end{figure*}

\begin{figure*}
    \centering
    
  \begin{Verbatim}[frame=single, framesep=2mm, fontsize=\small, breaklines=true]
Instruction: Write an accurate, engaging, and concise answer for the given question using only the provided search results (some of which might be irrelevant) and cite them properly by answering all the subquestions. Each subquestion should be answered. Use an unbiased and journalistic tone. Always cite for any factual claim. When citing several search results, use [1][2][3]. Cite at least one document and at most three documents in each sentence. If multiple documents support the sentence, only cite a minimum sufficient subset of the documents.
        
   \end{Verbatim}

    \caption{Instruction for answer generation of Blueprint}
    \label{fig:blueprint2}
\end{figure*}

\begin{figure*}
    \centering
    
  \begin{Verbatim}[frame=single, framesep=2mm, fontsize=\small, breaklines=true]
In this task, you will be given a question, and you should generate a query to find relevent documents to help generating the answer. You may be given some sentences that have been generated as context, you should try to find documents that could support another claim other than sentences generated but still relevent to the question. 

Given the original question: Who has the highest goals in world football?
Please generate one query to help find relevent documents, the query is:


Answer:
"Top goal scorer in world football 2021"
        
   \end{Verbatim}

    \caption{Retrieval query generation of Prompt self-RAG}
    \label{fig:selfrag1}
\end{figure*}

\begin{figure*}
    \centering
    
  \begin{Verbatim}[frame=single, framesep=2mm, fontsize=\small, breaklines=true]
Instruction: Write only a sentence as an accurate, engaging, and concise answer for the given question using only the provided search result. Use an unbiased and journalistic tone.

Question:Who has the highest goals in world football?

Prefix:Pelé holds the record for the highest number of goals in world football, with 1281 goals recognized by FIFA[1]. 

Document [4](Title:Wartan Ghazarian)goals (4 in World Cup qualifiers, 3 in Asian Cup qualifiers, 12 in friendlies). His record was later broken by Roda Antar, after Roda scored his 20th goal in 2018 FIFA World Cup qualification match against Laos. On 16 November 2008, during Round 6 of the Lebanese Football League, at the age of 39 years, Vartan scored his 130th goal in the Lebanese first division against Tadamon Tyre, becoming officially the highest all-time scorer in the history of Lebanese football. Some officials do not recognize the 12 goals he scored in the 2000–2001 season which was canceled. However, his remaining

Answer: 
        
   \end{Verbatim}
    \caption{Instruction for answer generation of Prompt self-RAG}
    \label{fig:selfrag2}
\end{figure*}

\begin{figure*}
    \centering
    
  \begin{Verbatim}[frame=single, framesep=2mm, fontsize=\small, breaklines=true]
Instruction: You will be presented with a snippet from documents. Write only a sentence as an accurate, engaging, and concise answer for the given question using only the provided snippets. Use an unbiased and journalistic tone.

Question:Who has the highest goals in world football?

Prefix:Pelé holds the record for the highest number of goals in world football, with 1281 goals recognized by FIFA[1]. 

Document [4](Title:Wartan Ghazarian)His record was later broken by Roda Antar, after Roda scored his 20th goal in 2018 FIFA World Cup qualification match against Laos.

Answer: 
        
   \end{Verbatim}
    \caption{Instruction for answer generation of Prompt self-RAG \snippet}
    \label{fig:selfragsnippet}
\end{figure*}
\end{document}